\documentclass{article}

\usepackage{PRIMEarxiv}

\usepackage[utf8]{inputenc} 
\usepackage[T1]{fontenc}    
\usepackage{hyperref}       
\usepackage{url}            
\usepackage{booktabs}       
\usepackage{amsfonts}       
\usepackage{nicefrac}       
\usepackage{microtype}      
\usepackage{lipsum}
\usepackage{fancyhdr}       
\usepackage{graphicx}       
\graphicspath{{media/}}     
\usepackage{xcolor}

\title{
A Moral- and Event- Centric Inspection of Gender Bias in Fairy Tales at A Large Scale
}
  
\author{
  Zhixuan Zhou \\
  University of Illinois at Urbana-Champaign \\
  \texttt{zz78@illinois.edu} \\
   \And
  Jiao Sun \\
  University of Southern California \\
  \texttt{jiaosun@usc.edu} \\
   \AND
  Jiaxin Pei \\
  University of Michigan \\
  \texttt{pedropei@umich.edu} \\
  \And
  Nanyun Peng \\
  University of California, Los Angeles \\
  \texttt{violetpeng@cs.ucla.edu} \\
   \And
  Jinjun Xiong \\
  University at Buffalo \\
  \texttt{jinjun@buffalo.edu} \\
}

\begin{document}
\maketitle

\begin{abstract}




    
    Fairy tales are a common resource for young children to learn a language or understand how a society works. However, gender bias, e.g., stereotypical gender roles, in this literature may cause harm and skew children's world view. Instead of decades of qualitative and manual analysis of gender bias in fairy tales, we computationally analyze gender bias in a fairy tale dataset containing 624 fairy tales from 7 different cultures. We specifically examine gender difference in terms of moral foundations, which are measures of human morality, and events, which reveal human activities associated with each character.  
    
    We find that the number of male characters is two times that of female characters, showing a disproportionate gender representation. Our analysis further reveal stereotypical portrayals of both male and female characters in terms of moral foundations and events.
    Female characters turn out more associated with care-, loyalty- and sanctity- related moral words, while male characters are more associated with fairness- and authority- related moral words. Female characters' events are often about emotion (e.g., weep), appearance (e.g., comb), household (e.g., bake), etc.; while male characters' events are more about profession (e.g., hunt), violence (e.g., destroy), justice (e.g., judge), etc. Gender bias in terms of moral foundations shows an obvious difference across cultures.
    For example, female characters are more associated with care and sanctity in high uncertainty-avoidance cultures which are less open to changes and unpredictability. Based on the results, we propose implications for children's literature and early literacy research.
    
    
\end{abstract}


\section{Introduction}
Gender bias is prevalent in news reports, social media, and books~\cite{gender:wikipedia:event, implicit:bias, mom:vlogging, gender:story}. Children are especially vulnerable to the influence of gender bias because they are still forming their values through various resources such as books. Fairy tales, a popular form of children books, can affect children's perspectives of the world. 
%
To explore the presence of gender bias in fairy tales, we conduct text mining on  
a fairy tale dataset \cite{dataset} of 624 fairy tales, with 161 of them containing culture information, i.e., 33 Norwegian, 28 Chinese, 24 Native American, 23 Swedish, 19 Japanese,  19 Scottish, and 15 Irish fairy tales. 


To the best of our knowledge, our work is one of the first to systematically and quantitatively analyze the presence of gender bias in fairy tales, through the lens of moral foundation~\cite{emfd} and event/event chain (a sequence of events)~\cite{eventplus}. 

\textbf{Moral Foundations.}
Moral foundations are measures of human morality. Prior literature has proposed five innate, universal moral foundation pairs, including care/harm (concerning intuitions of sympathy, compassion, and nurturance), fairness/cheating (concerning notions of rights and justice), loyalty/betrayal (concerning moral obligations of patriotism and ``us versus them'' thinking), authority/subversion (concerning traditions and maintaining social order), and sanctity/degradation (concerning moral disgust and spiritual concerns related to the body)~\cite{emfd}. Moral foundations have been extensively used as a lens to inspect social events and technologies in the social science literature, e.g., identifying moral dynamics shaping Twitter discussions around Black Lives Matter \cite{emfd:app:1, emfd:app:2, emfd:app:3}.

Inspired by this line of prior work, we measured gender bias in fairy tales by comparing the occurrences of moral words in both genders, as well as associated sentiments. Specifically, we used  eMFD\footnote{\url{https://github.com/medianeuroscience/emfdscore}} to score sentences associated with each character on the five dimensions of moral foundations, and compared moral foundation scores (both probability and sentiment) of different genders. We observed notable gender differences in our analysis. Female characters are associated with a larger ratio of moral words. Female characters are more associated with care-, loyalty- and sanctity- related moral words, while male characters are more associated with fairness- and authority- related moral words. In addition, female characters are more positively framed in all moral foundations than male characters, narrated on moral words of a higher sentiment score. 

\textbf{Events.}
While we observed gender differences in our moral foundation results, they are not always intuitive to understand. 
To help further unpack the gender differences, we compared events and event chains 
associated with both genders. Essentially, human activities can be seen as sequences of events, which are crucial to understanding gender portrayal.

Events have been used as a lens to understand gender bias in the Wikipedia corpus~\cite{gender:wikipedia:event}. We adopted a similar approach and enhanced it by extracting {\em event chains}, which helped us understand how a certain gender was consistently portrayed in a sequence of events. Specifically, we used EventPlus \cite{eventplus}, a temporal event understanding pipeline, to extract events and their temporal relations. We then used the metric of odds ratio (OR) \cite{or} 
to identify events, event types, and event chains specific to different genders. 

Using this event-centric analysis, we found events associated with female characters were more about emotion, appearance, care, loyalty, lower authority/power, household, and femininity, while events associated with male characters were more about profession, violence/masculinity, indulgence, travel/motion, justice, and higher authority/power. Considering event types tagged by EventPlus \cite{eventplus}, female characters were narrated more on life events, such as Life:Marry and Life:Be-Born, and male characters more on personnel/profession- and conflict- related events, such as Personnel:Start-Position, Personnel:Elect, Conflict:Attack and Conflict:Demonstrate. 
In our dataset, we saw such gender bias consistently appearing in event chains, e.g., women weep before they get married, while men kill before they get married.

\textbf{Cross-Cultural Analysis of Gender Bias.}
Different cultures have varying social norms and values, 
and gender bias is not likely to be the same. To understand the narratives of gender roles in different cultures, we adopted Hofstede's cultural dimensions theory to compare gender bias in different cultural contexts.

Gender bias in terms of moral foundations and events turned out to vary in different cultures. For example, male characters are more associated with fairness and authority in high power-distance cultures (those with a fixed power hierarchy), whereas female characters are more associated with these two moral dimensions in high individualism cultures and high indulgence cultures (those allowing the people to fulfill their human desires); female characters are more associated with care and sanctity, and more positively framed regarding care in high uncertainty-avoidance cultures (those less open to changes and unpredictability). 
Female characters in cultures with a long-term orientation (those having a pragmatic perspective, China and Japan in our case) are more feminine in terms of events.


Based on the findings above, we reflect on implications for future children's literature and early literacy research, e.g., adapting existing children's books and enabling critical reading of children, toward making our next generations more mindful of gender bias in books and more equipped to respond to gender bias in other contexts. 

This work makes three main contributions: (1) we provide an end-to-end pipeline to automatically identify gender bias in terms of moral foundations; (2) we mine events and event chains (a series of events in temporal order) associated with each character to explain the less interpretable moral foundation scores; and (3) we conduct a cross-cultural analysis of gender bias in fairy tales through the lens of Hofstede's cultural dimensions theory.

In the sections below, we will elaborate on dataset construction, gender bias analysis in terms of moral foundations and events/event chains, and the cross-culture analysis of gender bias. Related works and research/practical implications are discussed in the end.

\section{Dataset Construction}
Our data is curated from a fairy tale dataset \cite{dataset} consisting of 624 stories, which were crawled from Project Gutenberg\footnote{\url{https://www.gutenberg.org/}. Project Gutenberg is a volunteer effort to digitize and archive cultural works, and to encourage the creation and distribution of eBooks.}. Some of the fairy tales include the metadata of culture information, which are used for the cross-cultural analysis of gender bias. We follow the steps below to construct our dataset.

\subsection{Character Co-Reference}
As the first step, we need to recognize characters, as well as co-referent pronouns for each character (e.g., ``he'' refers to ``king'' in a certain fairy tale). We complete this task with BookNLP \cite{booknlp}, which uses a Bayesian mixed effects model to do co-reference. We have also explored other neural network-based models for person name recognition (e.g., \cite{NER:NN}), but given the satisfactory performance and time efficiency of BookNLP, it becomes our choice for character co-reference.

\subsection{Sentence Extraction and Grouping}
We extract sentences containing each character and concatenate them into a paragraph. Note that sentences associated with each character are not necessarily adjacent in the raw text.

\subsection{Gender Tagging}
The BookNLP output 
places co-referent characters after each pronoun, e.g., he (king), she (princess). We rely on BookNLP to decide the gender of each character. Specifically, we count male pronouns (i.e., he/him/his) and female pronouns (i.e., she/her) co-referent with each character. If there are more male pronouns than female pronouns co-referent with a certain character, we regard this character as male, and vice versa. A random tie-breaker is implemented. We randomly check the gender of 10 characters, confirming feasibility of this approach to gender identification.  

\subsection{Event Extraction}
We use EventPlus \cite{eventplus} to extract events associated with each character out of her concatenated sentences. While one sentence may contain multiple characters, we use a semantic role labeling (SRL) \cite{srl} tool\footnote{\url{https://demo.allennlp.org/semantic-role-labeling/semantic-role-labeling}} to filter out irrelevant events. For example, in the sentence ``Alice gave Bob's mom a book,'' Alice is the giver (ARG0), Bob's mom is the entity given to (ARG2), and the book is the thing given (ARG1).
We associate an event with a character only if she is in one of the three roles for the event. We lemmatize and move stop words from the events to ensure the quality of later event analysis.


\subsection{Sorting Events in Temporal Order}
The EventPlus output provides the temporal relations among events in each sentence. To sort events in multiple sentences associated with a character, we assume 
sentences of a character are in temporal order, which we acknowledge as a strong assumption. To gain more confidence about this assumption, we randomly sample 14 fairy tales (2 from each culture), and manually check how the stories are written. It is found that most fairy tales tell stories in typical narration, and pure flashbacks are not found, which might be attributed to the authors' consideration for children's limited reading capabilities. 

\subsection{Obtaining Moral Foundation Scores}
Each word in the eMFD dictionary has 5 probability scores and 5 sentiment scores (i.e., \{probability, sentiment\}$\times$\{care, fairness, loyalty, authority, sanctity\}). The probability score denotes the likelihood a word is associated with a certain moral foundation. For example, the word ``kill'' has a care probability of 0.4, meaning that there is a 40\% chance that a coder annotated a context containing the word ``kill'' with the care-harm foundation.
Sentiment score denotes the average sentiment of the context in which a word appeared. For example, the word ``kill'' has a ``care\_sent'' of -0.69, meaning that ``care-harm'' contexts containing the word ``kill" had an average, negative VADER sentiment score \cite{vader} of -0.69. See more details in the eMFD tutorial\footnote{\url{https://github.com/medianeuroscience/emfdscore/blob/master/eMFDscore_Tutorial.ipynb}}.

We use eMFD to obtain moral foundation scores for sentences associated with each character, including 5 probability scores, 5 sentiment scores, and the ratio of moral words and non-moral words. 

Note that we choose not to run 
lemmatization for the input of eMFD 
since the eMFD dictionary contains verbs in different tenses, with different moral foundation scores. For example, the words ``bring'' and ``brought'' differ in moral foundation scores. 

\subsection{Dataset Description}    
In the end, our dataset contains such information as fairy tale title, culture (for those without culture information, we mark the culture information as ``unknown''), character (a fairy tale can contain multiple characters, thus each character is presented in a row), gender of character (2,125 female characters, 4,405 male characters), count of appearance of each character, concatenated sentences associated with each character, eMFD scores (11 scores in total: 5 probabilities, 5 sentiments, 1 moral vs. non-moral words ratio), and extracted events in temporal order.

\section{Different Moral Representations Across Genders}
To examine the gender difference in moral foundations, we calculate the average eMFD scores (11 in total) for female and male characters, respectively, and confirm the between-gender differences using a t-test. 

\subsection{Mentions of Moral Foundations}
The gender difference regarding mentions of moral foundations is confirmed by t-test on probability scores (see Table~\ref{tab:emfd:all}). In general, female characters are more associated with care, loyalty, and sanctity. Male characters are more associated with fairness and authority. From moral\_nonmoral ratio (ratio of moral words to non-moral words), we can see that female characters are more associated with moral words, suggesting more moral narratives around female characters in fairy tales.

\subsection{Moral Foundation Sentiments}
The gender difference regarding moral foundation sentiment is also confirmed by the t-test on sentiment scores (see Table~\ref{tab:emfd:all}). A higher sentiment score means being more positively framed regarding a moral foundation dimension in fairy tales. It is found that female characters are more positively framed in moral foundations than male characters in the care and fairness dimensions. There are no significant differences between male and female in the loyalty, authority, and sanctity dimensions. 



\subsection{Events-Only Analysis}
To see how events alone contribute to gender bias, we feed only extracted events in the sentences into eMFD, and compare the results between male and female characters. Results are shown in Table~\ref{tab:emfd:event}.

Gender difference in terms of moral foundation sentiments still holds: female characters are framed more positively regarding moral foundations. However, gender difference in terms of moral foundation probabilities becomes insignificant after removing context, except that female characters are more associated with the loyalty dimension, and that a larger ratio of moral words exist in the narrative of female characters. This can be attributed to the fact that in the eMFD dictionary, there are not only events (mostly verbs), but also nouns and adjectives, all of which are important to understanding moral foundations in fairy tales.

\begin{table*}
\centering
\begin{tabular}{lcccccc}
\hline
\textbf{Attributes (Avg)} & \textbf{Male} & \textbf{Female} & \textbf{Ratio (M/F)} & \textbf{\textit{p}-value} & \textbf{\textit{t}-statistic} & \textbf{More mention/moral}\\
\hline
\textbf{Care\_p} & 0.1090 & 0.1109 & 0.9831 & <0.05 & -3.2 & female(frequent)\\
Care\_sent & -0.0896 & -0.0851 & 1.0519 & <0.05 & -2.0 & female(moral)\\
\hline
Fairness\_p & 0.0950 & 0.0940 & 1.0116 & <0.05 & 2.4 & male(frequent) \\
\textbf{Fairness\_sent} & -0.0294 & -0.0126 & 2.3272 & <0.05 & -7.9 & female(moral)\\
\hline
\textbf{Loyalty\_p} & 0.0919 & 0.0935 & 0.9829 & <0.05 & -3.8 & female(frequent)\\
Loyalty\_sent & -0.0045 & -0.0012 & 3.7753 & >0.05 & -1.6 & not significant(moral)\\
\hline
\textbf{Authority\_p} & 0.0910 & 0.0881 & 1.0325 & <0.05 & 6.6 & male(frequent)\\
Authority\_sent & -0.0174 & -0.0175 & 0.9942 & >0.05 & 0.0 & not significant(moral)\\
\hline
\textbf{Sanctity\_p} & 0.0859 & 0.0891 & 0.9650 & <0.05 & -7.3 & female(frequent)\\
Sanctity\_sent & -0.0588 & -0.0596 & 0.9854 & >0.05 & 0.4 & not significant(moral)\\
\hline
Moral\_nonmoral\_ratio & 1.4941 & 1.5663 & 0.9539 & <0.05 & -2.6 & female(frequent)\\\hline
\end{tabular}
\caption{Gender difference in moral foundation scores (raw text). p-probability, 
sent-sentiment. The last column denotes the gender which is mentioned more on moral foundations (according to probability score), or does better morally (according to sentiment score). There are 4,405 male characters, and 2,125 female characters in our dataset.}
\label{tab:emfd:all}
\end{table*}

\begin{table*}
\centering
\begin{tabular}{lcccccc}
\hline
\textbf{Attributes (Avg)} & \textbf{Male} & \textbf{Female} & \textbf{Ratio (M/F)} & \textbf{\textit{p}-value} & \textbf{\textit{t}-statistic} & \textbf{More mention/moral}\\
\hline
Care\_p & 0.0995 & 0.0999 & 0.9963 & >0.05 & -0.4 & not significant(frequent)\\
Care\_sent & -0.1164 & -0.1088 & 1.0700 & <0.05 & -2.7 & female(moral)\\
\hline
Fairness\_p & 0.0854 & 0.0864 & 0.9882 & >0.05 & -1.5 & not significant(frequent) \\
Fairness\_sent & -0.0720 & -0.0648 & 1.1107 & <0.05 & -2.4 & female(moral)\\
\hline
Loyalty\_p & 0.0924 & 0.0941 & 0.9816 & <0.05 & -2.7 & female(frequent)\\
\textbf{Loyalty\_sent} & -0.0336 & -0.0259 & 1.2972 & <0.05 & -3.1 & female(moral)\\
\hline
Authority\_p & 0.0883 & 0.0886 & 0.9965 & >0.05 & -0.4 & not significant(frequent)\\
Authority\_sent & -0.0444 & -0.0419 & 1.0597 & >0.05 & -0.9 & not significant(moral)\\
\hline
Sanctity\_p & 0.0758 & 0.0767 & 0.9885 & >0.05 & -1.5 & not significant(frequent)\\
Sanctity\_sent & -0.0654 & -0.0618 & 1.0571 & >0.05 & -1.3 & not significant(moral)\\
\hline
Moral\_nonmoral\_ratio & 2.6172 & 2.7735 & 0.9437 & <0.05 & -2.7 & female(frequent)\\\hline
\end{tabular}
\caption{Gender difference in moral foundation scores (event trigger words). p-probability, sent-sentiment. The last column denotes the gender which is mentioned more on moral foundations (according to probability score), or does better morally (according to sentiment score). There are 4,405 male characters, and 2,125 female characters in our dataset.}
\label{tab:emfd:event}
\end{table*}

\section{Different Events Across Genders}

With EventPlus, we obtain two dictionaries of events, $\mathcal{E}^{m}$ for male and $\mathcal{E}^{f}$ for female. The dictionaries have events as keys and their corresponding occurrence frequencies as values. We focus on events with distinct occurrences in the narratives of male and female characters and neglect those with a similar occurrence in both genders. To find the events with a large frequency difference between female and male characters, for each event $e_{n}$, we calculate its odds ratio \cite{or}, i.e., the odds of having it in the male event list divided by the odds of having it in the female event list: 
\begin{equation}
\frac{\mathcal{E}^{m}\left(e_{n}\right)}{\sum_{e_{i}^{m} \neq e_{n} \atop i \in[1, \ldots, M]}^{i} \mathcal{E}^{m}\left(e_{i}^{m}\right)} / \frac{\mathcal{E}^{f}\left(e_{n}\right)}{\sum_{e_{j}^{f} \neq e_{n} \atop j \in[1, \ldots, F]}^{j} \mathcal{E}^{f}\left(e_{j}^{f}\right)}
\end{equation}
By this equation, the larger the odds ratio is, the more likely an event will occur in male than female characters. We apply a similar odds ratio analysis to event types. 
Events before/after certain events are also extracted to identify event chains that are prominent in the portrayal of different genders. We list the top 20 events and event types for both genders in Table \ref{tab:or:event}, and events before/after certain selected events in Table \ref{tab:or:neighbor}. 


\subsection{Events}

\begin{table*}
\centering
\begin{tabular}{p{30mm}p{10mm}p{100mm}}
\hline
\textbf{Event} & \textbf{Gender} & \textbf{Top-20}\\ 
\hline
Event word (agency and power) & Male & arise (pos,-), shoot (pos,agent), hit (pos,agent), leap (pos,-), chop (pos,agent), land (pos,-), stick (pos,-), describe (pos,equal), judge (pos,agent), entertain (pos,theme), descend (pos,-), cross (pos,agent), hunt (pos,agent), thrust (pos,agent), disturb (pos,equal), borrow (pos,agent), destroy (pos,agent), appoint (pos,agent), praise (pos,equal), bethink (-,-)\\
& Female & spin (pos,agent), comb (pos,agent), bake (pos,agent), reveal (pos,agent), dry (-,-), soothe (pos,equal), starve (pos,agent), dive (pos,-), enable (pos,agent), lament (-,-), adorn (pos, theme), quarrel (-,-), foretell (-,-), foresee (-,-), clean (pos,agent), blush (neg,-), perish (-,-), stray (pos,-), betray (pos,agent), kindle (-,-)\\
\hline
Event type & Male & Justice:Release-Parole, Personnel:Start-Position, Justice:Execute, Personnel:Elect, Justice:Arrest-Jail, Conflict:Attack, Transaction:Transfer-Money, Conflict:Demonstrate, Personnel:End-Position, Contact:Meet, Justice:Sentence, Business:Start-Org, Life:Die, Transaction:Transfer-Ownership, Contact:Phone-Write, Life:Injure, Movement:Transport, Personnel:Nominate, Life:Be-Born, Justice:Charge-Indict\\
& Female & Justice:Sue, Life:Marry, Justice:Charge-Indict, Life:Be-Born, Personnel:Nominate, Movement:Transport, Life:Injure, Contact:Phone-Write, Transaction:Transfer-Ownership, Life:Die, Business:Start-Org, Justice:Sentence, Contact:Meet, Personnel:End-Position, Conflict:Demonstrate, Transaction:Transfer-Money, Conflict:Attack, Justice:Arrest-Jail, Personnel:Elect, Justice:Execute\\
\hline
\end{tabular}
\caption{Top 20 events and event types for male and female characters (selected by odds ratio).}
\label{tab:or:event}
\end{table*}

\begin{table*}
\centering
\begin{tabular}{p{30mm}p{10mm}p{50mm}}
\hline
\textbf{Event} & \textbf{Gender} & \textbf{Top-5}\\ 
\hline
Before marry & Male & know, say, get, come, read\\
& Female & give, see, cry, wish, make\\

After marry & Male & send, reply, live, say, find\\
& Female & come, read, have, ask, saw\\
\hline
Before say & Male & stop, eat, show, rise, point\\
& Female & wish, travel, rub, build, smile\\

After say & Male & ride, eat, follow, command, listen\\
& Female & kiss, appear, creep, comfort, hand\\
\hline
Before cry & Male & call, find, say, answer, saw\\
& Female & have, tell, fell, open, turn\\

After cry & Male & reply, come, have, go, throw\\
& Female & tell, let, marry, make, leave\\
\hline
Before beg & Male & ask, say, tell, have, answer\\
& Female & come, take, beg, see, knock\\

After beg & Male & leave, stay, let, be, answer\\
& Female & consider, give, beg, ask, take\\
\hline
\end{tabular}
\caption{Top 5 neighboring events before and after selected common life events (selected by odds ratio).}
\label{tab:or:neighbor}
\end{table*}

We examine top events for both genders in terms of odds ratio. Events in the portrayal of female characters are more about emotion (e.g., lament), appearance (e.g., comb), care (e.g., soothe), loyalty (e.g., betray), lower authority/power (e.g., starve), household (e.g., bake, adorn), and femininity (e.g., blush). On the other hand, male characters are more associated with events about profession (e.g., hunt, appoint), violence/masculinity (e.g., shoot, hit, chop, stick, thrust, destroy), indulgence (e.g., entertain), travel/motion (e.g., arise, leap, cross), justice (judge), and higher authority/power (e.g., praise).

\subsection{Event Types}
Event types are high-level summaries of events associated with different genders. 
While the taxonomy of event types provided by EventPlus is drawn from data in a more modern setting (e.g., Contact:Phone-Write), gender difference can still be seen by calculating the odds ratio of event types. See event type results in Table~\ref{tab:or:event}.

Notably, female characters are narrated more on life events, which can be categorized into such event types as Life:Marry, Life:Be-Born, Life:Injure and Life:Die. Male characters are narrated more on personnel/profession- and conflict- related events, which can be categorized into such event types as Personnel:Start-Position, Personnel:Elect, Personnel:End-Position, Conflict:Attack, and Conflict:Demonstrate. As we can see, the portrayal of female characters is often limited to their personal life, while male characters are portrayed as individuals with professions and power.



\subsection{Event Chains}
\label{bow}
We examine events before/after certain events for male and female characters, respectively, to have a holistic view of the portrayals of different genders. The events which we choose to look ahead and back are either typical female/male events selected by odds ratio (e.g., cry for female characters), or common life events \cite{gender:wikipedia:event} of the authors' selection (e.g., marry) which appear frequently in both female and male characters. 
By calculating the odds ratio of events before/after these events, apparent gender differences can be seen. Below are some examples.

``Marry'' is a power-equal and agency-neutral event, which always includes two genders (other gender identities are rare in ancient-time fairy tales). 
Interestingly, women weep, and men kill before their marriage, showing a clear discrepancy in masculinity/femininity. Other events associated with women before ``marry'' include care, dance, love, and beg, and events associated with men include cut and ride. It's worth noticing that the event ``marry'' appears 195 times in female characters, and 223 times in male characters. Given that there are 107.29\% more male characters than female characters, female characters are more frequently associated with ``marry.''


Before women ``cry'', they would fling, frighten, sigh, implore and sob, consistently portrayed as emotional and weak. Before men ``cry'', they command, sing, threaten, win and order, portrayed as much more powerful and with more agency than women. After ``cry'', women wail, comfort, and implore, while men drink and scold.


As we can see in these examples, female and male characters are associated with different events before/after the same events. Women are consistently portrayed as caring, emotional, and weak in power and agency, while men are consistently portrayed as professional, violent, and strong in power and agency.

\subsection{Using Events to Interpret Moral Foundations}
While moral foundation scores are not informative and intuitive enough to understand the difference between portrayals of male and female characters, the extracted events can speak for moral foundations. For example, the frequent occurrence of care- and household- related events in female characters, and the absence of these events in male characters explain why female characters are more associated with the care dimension in the moral foundation analysis. Similarly, the relatively frequent occurrence of justice-related events in male characters echoes with the finding that male characters are more associated with the fairness dimension regarding moral foundations. While female characters are found to be more positively framed in all moral foundations than male characters, events associated with male characters are indeed less moral, with the prevalence of violence-related events such as shoot and destroy.

 
\section{Are Gender Bias the Same in Different Cultures?}
Since culture information is available for some fairy tales, we are able to examine gender bias in different cultural contexts, namely, (1) male/female ratio of moral foundation probability scores, (2) male-female difference of moral foundation sentiment scores, and (3) femininity of female characters (ratio of female events to male events selected by odds ratio) and masculinity of male characters (ratio of male events to female events). 

We quantify cultural backgrounds using Hofstede's cultural dimensions theory \cite{culture}, which concludes six dimensions of national culture. In this theory, culture is defined as the collective mental programming of the human mind which distinguishes one group of people from another. The six indices describing culture include:
\begin{itemize}
    \item Power distance index (PDI): PDI reflects ``the extent to which the less powerful members of organizations and institutions accept and expect that power is distributed unequally.'' A higher degree of PDI indicates that hierarchy is clearly established and executed in society, without doubt or reason.
    \item Individualism (IDV) vs. collectivism: IDV reflects the ``degree to which people in a society are integrated into groups.'' Individualistic societies emphasize the ``I'' versus the ``we'', and have loose ties that often only relate an individual to his/her immediate family.
    \item Uncertainty avoidance (UAI): UAI reflects ``a society's tolerance for ambiguity.'' A lower degree in this index shows more acceptance of differing thoughts or ideas, which often means fewer regulations, and a more free-flowing environment.
    \item Masculinity (MAS) vs. femininity: masculinity reflects ``a preference in society for achievement, heroism, assertiveness and material rewards for success,'' while femininity reflects ``a preference for cooperation, modesty, caring for the weak and quality of life.'' 
    \item Long-term orientation (LTO) vs. short-term orientation: LTO can be thought of as the degree to which a culture has a pragmatic perspective that is focused on the future. Societies with a long-term orientation are willing to sacrifice present-day comforts for the sake of future rewards. Short-term orientation, on the other hand, can be thought of as the degree to which a society has a perspective that is more tradition-oriented.
    \item Indulgence (IND) vs. restraint: IND reflects the degree of freedom that societal norms give to citizens in fulfilling their human desires.
\end{itemize}

We collect the six culture indices for the seven cultures in our analysis from Hofstede Insights\footnote{\url{https://www.hofstede-insights.com/country-comparison/}} (listed in Table ~\ref{culture}), and calculate Pearson’s correlation between the six cultural indices and average gender bias indices defined by us for each culture. We associate Scottish fairy tales with the United Kingdom, and Native American fairy tales with the United States, which is hardly a good practice in the children's literature field, and we acknowledge this as a limitation. Our approach has been inspired by another study on cross-culture analysis of emoji use \cite{emoji}. 

\begin{table*}
\centering
\begin{tabular}{p{30mm}p{10mm}p{10mm}p{10mm}p{10mm}p{10mm}p{10mm}}
\hline
\textbf{Index/Country} & \textbf{PDI} & \textbf{IDV} & \textbf{MAS} & \textbf{UAI} & \textbf{LTO} & \textbf{IND}\\
\hline
United States & 40 & 91 & 62 & 46 & 26 & 68\\
China & 80 & 20 & 66 & 30 & 87 & 24\\
Ireland & 28 & 70 & 68 & 35 & 24 & 65\\
Japan & 54 & 46 & 95 & 92 & 88 & 42\\
Norway & 31 & 69 & 8 & 50 & 35 & 55\\
United Kingdom  & 35 & 89 & 66 & 35 & 51 & 69\\
Sweden & 31 & 71 & 5 & 29 & 53 & 78\\
\hline
\end{tabular}
\caption{Culture indices for the 7 cultures in our analysis.}
\label{culture}
\end{table*}

\subsection{Cultural Difference Through the Lens of Moral Foundations}
Considering male/female ratio of moral foundation probabilities, we have the following results:
\begin{itemize}
    \item Male characters are more associated with fairness and authority in higher power distance cultures;
    \item Female characters are more associated with fairness and authority in higher individualism cultures and higher indulgence cultures; 
    \item Female characters are more associated with care and sanctity in higher uncertainty-avoidance cultures.
\end{itemize}
Considering male-female difference of moral foundation sentiments, we find that female characters are more positively framed regarding care in higher uncertainty-avoidance cultures.

We can see that more gender stereotypical portrayals are seen in higher power distance cultures which believe there is nothing wrong with inequality and everyone has specific positions, and higher uncertainty-avoidance cultures where there is a greater emphasis on rules, structure, order, and predictability. On the other hand, less gender stereotypical portrayals are seen in higher individualism cultures where individuality and individual rights are paramount, and higher indulgence cultures where having fun and fulfilling desires are considered natural.

\cite{moral:sex} is another work discussing gender differences in moral judgments. They find that women consistently score higher than men on care, fairness, and purity; gender differences in moral judgments are larger in individualist and gender-equal societies with more flexible social norms. They have a different perspective than ours: they regard gender difference in moral judgments as a good thing since the moral judgments are self-perceived by individuals (collected with surveys); whereas we regard gender difference in moral foundations as gender bias/stereotypes, since it reflects different gender portrayals in fairy tales.

\subsection{Cultural Difference Through the Lens of Events}


Femininity score (ratio of female events to male events) of female characters is positively correlated with LTO, i.e., female characters in cultures with a long-term orientation (China and Japan in our data) are more feminine. However, the femininity of female characters and masculinity of male characters are not significantly correlated to other culture indices, which can be attributed to the limited size of the corpus, and the low number of events selected by odds ratio.





\section{Related Work}
NLP research hardly touched the pressing ethical issues of gender and race until very recently. Field at al. \cite{racial:NLP} conducted a literature survey on race, racism, and anti-racism in NLP. They found that NLP research on race was restricted to specific tasks and applications, while the race issue in many common NLP applications had not been examined enough, such as machine translation, summarization, and question answering. Similarly, Birhane et al. \cite{value:ML} conducted a qualitative analysis of 100 most cited papers from NeurIPS and ICML, revealing that values related to user rights and stated in ethical principles rarely occurred in the papers. Other moral values such as autonomy, justice, and respect for people were also largely absent.

\subsection{Uncovering Gender Bias in Different Corpus}
There are several papers utilizing NLP techniques to identify gender bias in different media, including textbooks, model-generated stories, and Wikipedia. Sun et al. \cite{gender:mitigation} systematically reviewed studies on recognizing and mitigating gender bias in NLP and proposed future directions such as mitigating gender bias in languages beyond English.

Gender bias has often been found in NLP models. Huang et al. \cite{implicit:bias} used a commonsense reasoning engine to uncover implicit gender bias associated with protagonists in model-generated stories, showing that female characters' portrayal was centered around appearance, while male figures' portrayal focused on intellect. Lucy and Bamman \cite{gender:story} used topic modeling and lexicon-based word similarity (appearance, intellect, and power) to examine stories generated by GPT-3 using a prompt, and found they exhibited many known gender stereotypes. For example, feminine characters were more associated with family and appearance, and described as less powerful than masculine characters, even when associated with high-power verbs in a prompt. 

Gender bias and stereotypes are also widely present in life scenarios which people frequently interact with, e.g., Wikipedia \cite{gender:wikipedia:event}, textbook \cite{textbook}, social media \cite{gender:weibo}, greeting cards \cite{greeting:card}, real-world AI applications \cite{gender:incident}, etc. Lucy et al. conducted a content analysis on Texas U.S. history textbooks and found the most common famous figures were nearly all White men, and women tended to be discussed in the contexts of work and the home \cite{textbook}. Sun and Peng \cite{gender:wikipedia:event} presented the first event-centric study of gender bias in a Wikipedia corpus by detecting events with an event detection model. They found Wikipedia pages tended to mix personal life events with professional events for females but not for males, and thus called for the awareness of the Wikipedia community to mind the implicit bias contributors carry. Similarly, Sun et al. uncovered a wide range of gender stereotypes in greeting cards via topic modeling, odds ratio, and Word Embedding Association Test (WEAT) \cite{greeting:card}. In a more holistic effort, word embedding was applied to uncover gender stereotypes from 7,226 books, 6,087 movie synopses, and 1,109 movie scripts. It turned out that the lives of males were adventure-oriented, whereas the lives of females were romantic-relationship oriented \cite{complex}.

In addition to content analysis and text mining techniques mentioned above, new approaches are being invented to uncover gender bias. For example, Field and Tsvetkov \cite{gender:discovery} discovered implicit gender bias in an unsupervised manner: if a classifier predicts the gender of the addressee with high confidence based on the text directed to them, they hypothesize that the text is likely to contain gender bias. Sap et al. introduced connotation frames of power and agency to model how different levels of power and agency were implicitly projected on actors in modern films through their actions \cite{movie}. The connotation frames confirmed evidence of imbalances in gender portrayal in movies. Their approach intrinsically looks at the nuanced differences regarding verb use in different genders. However, when we use their dictionary to annotate events regarding agency (positive, negative, equal) and power (agent, theme, equal), we find that even agency-positive and power-agent verbs can contain gender bias. For example, the frequent occurrence of such events as ``bake'' (agency-positive, power-agent) in female characters puts a stereotypical emphasis on women's household role. See our annotations in Table~\ref{tab:or:event}. Notably, our work is the first attempt of combining moral foundations and events/event chains in the identification of gender bias.

\subsection{Gender Bias in Children's Materials}
Cinderella-style fairy tales are known to harmfully reinforce restrictive images of girlhood and womanhood. In an early research, responses from children through group discussion, drawing pictures, and writing stories revealed that few girls identified with the pretty princess image \cite{image}. Unfortunately, teachers are not always concerned with gender stereotypes in fairy tales. Instead, they were reported to have a set of stereotyped expectations about how females and males were portrayed \cite{teacher}. Thus more efforts are needed to uncover and educate the public about gender bias in children's literature.

Qualitative efforts toward uncovering gender bias in children's books were seen back in 2002, when Gooden and Gooden \cite{gender:children} conducted a qualitative analysis of children’s picture books published between 1995 and 1999. Male adults tended to be illustrated in more roles (25 in total) than female adults (14 in total). Most of the women's roles were traditional ones, such as mother and washerwoman. Males were seldom seen caring for the children or doing grocery shopping, and never seen doing household chores. More recently, in 2013, Cekiso \cite{gender:fairy} conducted a qualitative discourse analysis on 2 fairy tales (Cinderella, and Snow White and the Seven Dwarfs) to deconstruct the texts concerning how female and male characters were portrayed. It was found that both genders were frequently presented in stereotypical terms: while the females were portrayed as submissive and dependent on men to rescue them, males were portrayed as having power, bravery, strength, and wit. 

Coats has discussed whether inaccurate or biased representations in children's books are harmful for children, and how we might invite children to rewrite or participate \cite{coats}. However, there is currently a lack of best practices of discovering and mitigating such misrepresentations at a large scale. A concurrent work with us measures gender portrayal of central domains of social life in highly influential children's books using word embedding \cite{anonymous}. Different from this work, we approach gender portrayal in children's books from an event- and moral- centric point of view. 


\subsection{eMFD and EventPlus}
Our analysis pipeline relies on eMFD \cite{emfd} to score sentences on five moral foundation dimensions, and EventPlus to extract events and their temporal relations \cite{eventplus}. 

eMFD (extended Moral Foundations Dictionary) is a dictionary-based tool for extracting moral content from textual corpora, which is constructed from text annotations generated by a large sample of human coders \cite{emfd}. Much work has been devoted to utilizing eMFD to inspect moral foundations embedded in social events and technologies \cite{emfd:app:1, emfd:app:2, emfd:app:3}. For example, Priniski et al. studied moral dynamics shaping Black Lives Matter Twitter discussions by analyzing Tweets geo-located to Los Angeles \cite{emfd:app:1}. Mutlu et al. presented the differences in the strength and patterns of moral rhetoric between human and bot-generated content on Twitter \cite{emfd:app:2}.

EventPlus is a temporal event understanding pipeline that integrates event trigger and type detection, event
argument detection, and event duration and temporal relation extraction \cite{eventplus}. It is a convenient tool for users to obtain annotations about events and their temporal information. Sun et al. utilized this tool to detect events in Wikipedia profiles and further uncover gender bias in this corpus \cite{gender:wikipedia:event}. 

\section{Discussion}
\subsection{Recap of Findings}
Apparent gender differences are found in the portrayal of male and female characters in fairy tales, regarding moral foundations and events/event chains. 

In the first place, there are 107.29\% more male characters than female characters. While we did not try to distinguish protagonists and supporting roles in the current analysis, having nearly 2 times more male characters than female characters is a misrepresentation of gender, let alone the lack of non-binary characters in the fairy tales. 

Through the lens of moral foundations, we find that female characters are more associated with care, loyalty, and sanctity, while male characters are more associated with fairness and authority. Female characters are more associated with moral words and are more positively framed in all moral foundations than male characters.

Through the lens of events, female characters are more associated with emotion, appearance, care, loyalty, lower authority/power, household, and femininity, while male characters are more associated with profession,  violence/masculinity, indulgence, travel/motion, justice, and higher authority/power. Female characters are narrated more on life events such as Life:Marry and Life:Be-Born; male characters are narrated more on personnel/profession- and conflict- related events, such as Personnel:Start-Position, Personnel:Elect, Conflict:Attack, and Conflict:Demonstrate.
By examining events before/after selected events, we find that female characters are consistently portrayed as caring, emotional, and weak in power and agency, while male characters are consistently portrayed as professional, violent, and strong in power and agency, showing a stereotypical gender portrayal.

The moral foundation and event results consistently uncover gender bias in fairy tales. Moreover, events are found to be an ideal lens to interpret moral foundation scores, which are less intuitive.

Gender bias varies in different cultures. In terms of moral foundation mentions, males are more associated with fairness and authority in higher power distance cultures; females are more associated with fairness and authority in higher individualism cultures and higher indulgence cultures; females are more associated with care and sanctity in higher uncertainty-avoidance cultures. In terms of moral foundation sentiments, females are more positively framed regarding care in higher uncertainty-avoidance cultures. A difference of femininity of female characters could also be seen across cultures through the lens of events.

\subsection{A Moral- and Event (Chain)- Centric Inspection of Gender Bias in Fairy Tales}

We are the first to combine moral foundations and events/event chains to uncover gender bias in fairy tales. Our moral-centric inspection of gender bias has been inspired by a line of work applying eMFD to understand social dynamics \cite{emfd:app:1, emfd:app:2}, and our event-centric inspection of gender bias has been inspired by a prior work on Wikipedia bias \cite{gender:wikipedia:event}. We extend the event analysis in \cite{gender:wikipedia:event} by examining event chains, i.e., sequences of events, in the portrayal of different genders, in addition to isolated events. A consistent portrayal pattern was identified using this approach, i.e., women are consistently portrayed as feminine and weak, and men are consistently portrayed as masculine and powerful (or brutal).    

Due to the lack of explainability afforded by the Moral Foundation lexicon, most prior work applying it to a new corpus (e.g., \cite{emfd:app:2}) can only report scores in each moral dimension without further interpretation. We find events and event chains to be an ideal lens to interpret moral foundation scores, which together uncover gender bias in children's literature, and can be applied to other media such as various book genres and online forums as well.

\subsection{Implications for Children's Literature and Early Literacy Research}

\textbf{Adapting classical fairy tales.}
It is nearly a cliche that ``increased effort is needed on the part of publishers and authors to provide children with literature that more closely parallels the roles of males and females in contemporary society \cite{sex:role}.'' Still, classical children's literature such as Grimm's Fairy Tales are widely read by children. They may contain centuries of human bias, teaching children stereotypical gender roles. 

Sporatic efforts have been devoted to adapting classical fairy tales into something more suitable for the modern society. For instance, the book Gender Swapped Fairy Tales \cite{gender:swap} by Karrie Fransman and Jonathan Plackett swapped genders of the characters, resulting in kings pricking their fingers as they sew, wolves wearing heels, and princesses racing to rescue sleeping princes. 

While such storytelling may not truly reflect the reality, more fine-grained adaptions can be made. For instance, given the large volume of gender stereotypes carried by events as found by our analysis, one can adjust the number of masculine or feminine events for the protagonists. In addition, in the adapted fairy tales, female characters should be assigned more roles, both traditional and non-traditional ones \cite{gender:children}. 

\textbf{Enabling active and critical reading of children.}
A lack of awareness of gender stereotypes in children's literature from both children and parents/teachers is yet another challenge for protecting children from gender bias \cite{teacher}. Though research has been devoted to uncovering gender stereotypes in children's literature, both qualitatively \cite{gender:fairy} and quantitatively \cite{anonymous}, the results are rarely conveyed to the public.

Toward raising children's awareness of gender stereotypes in fairy tales, one can build an end-to-end gender bias detector searching and presenting stereotypical events among other linguistic cues. Along with the highlighted text carrying stereotypical gender portrayals, children will also be presented a brief discussion of more modern gender roles (e.g., ``nowadays, women are no longer passive as narrated here, but instead equal decision makers''). 

Such a tool encouraging active and critical reading may help children figure out how gender bias works, by discovering and critiquing normative gender roles in fairy tales. That way, they are ready to see the same patterns in life, and when people express gender bias in other contexts, they are equipped to respond. 


\subsection{Limitations and Future Work}
There are several limitations of our work, as well as a wide space for future work. 

Firstly, we use BookNLP to identify the gender of each character, which treats gender as a binary variable (male and female). It could be interesting to examine how fairy tales portray characters of other gender identities, which we leave as future work. 

Secondly, out of 624 fairy tales, only 161 have culture information. With more fairy tales including culture information, more cultural differences in gender bias could have been uncovered. Also, we acknowledge that based on a sample of 7 cultures, our results may not generalize under Hofstede's culture theory framework. Besides, publication year of the fairy tales is missing in our data. Future work might want to control for publication year because cultural values can evolve over time.

Thirdly, when extracting event chains, we assume sentences of a character are always in temporal order. We acknowledge this as a strong assumption. We manually check 14 fairy tales and find most of the stories to be in typical narration. However, given the efforts required to manually check the narrative manner of all fairy tales, our confidence is only built upon this small sample. 

Fourthly, while the fairy tales we analyze originated in different cultures, they are analyzed in their English version. Gender bias can be introduced or mitigated in the translation process. An effort to examine gender bias in their original language is encouraged.  


Lastly, it would be interesting to compare children's literature with adult literature regarding gender bias/stereotypes. For example, characters in children's books might be more simplified (men are bread winners, women are home carers; men are brutal, women are kind) than in adult books.


\section{Conclusion}
Children are most easily affected by gender bias and stereotypes, thus it is important to analyze whether and how gender bias exists in children's literature. We approach this by understanding the different portrayal of male and female characters in fairy tales. Different moral portrayals across genders were identified: female characters are more associated with care-, loyalty- and sanctity- related moral words, while male characters are more associated with fairness- and authority- related moral words; female characters are more positively framed morally. Male and female characters are also found to be portrayed differently in terms of events and event chains, echoing with social stereotypes long associated with the two genders. A cross-culture analysis reveals the different existence of gender bias in different cultures. Based on the findings, we reflect on implications for future children's literature and early literacy research. Our work sheds light on understanding gender bias in children's literature at a large scale, and raising people's awareness of gender bias.

\section*{Acknowledgments}
We would like to thank Dakuo Wang and Yufang Hou from IBM, and Yang Wang from the University of Illinois at Urbana-Champaign for their help in designing and implementing the analysis. We further thank Elizabeth Hoiem from the University of Illinois at Urbana-Champaign for her invaluable inputs which helped the authors better situate the results in children's literature research.

\bibliographystyle{unsrt}  
\bibliography{references}  

\end{document}